\pdfoutput=1

\documentclass[11pt]{article}

\usepackage[]{acl}

\usepackage{times}
\usepackage{latexsym}
\usepackage{amsfonts} 
\usepackage{amsmath}
\DeclareMathOperator*{\argmax}{argmax}
\usepackage{mathtools}
\usepackage{multirow}
\usepackage{enumitem} 
\usepackage{graphicx}
\usepackage{lipsum}
\usepackage{url}
\usepackage{caption}
\usepackage{subcaption}
\usepackage[T1]{fontenc}

\usepackage[utf8]{inputenc}

\usepackage{microtype}

\title{KenMeSH: Knowledge-enhanced End-to-end Biomedical Text Labelling}

\author{Xindi Wang$^{1,3}$, Robert E. Mercer$^{1,3}$, and Frank Rudzicz$^{2,3,4}$\\

$^1$Department of Computer Science, University of Western Ontario, London, Ontario, Canada\\
$^2$Department of Computer Science, University of Toronto, Toronto, Ontario, Canada \\
$^3$Vector Institute for Artificial Intelligence, Toronto, Ontario, Canada\\
$^4$ Unity Health Toronto, Toronto, Ontario, Canada\\
\texttt{xwang842@uwo.ca, mercer@csd.uwo.ca, frank@cs.toronto.edu}}

\begin{document}
\maketitle
\begin{abstract}
Currently, Medical Subject Headings (MeSH) are manually assigned to every biomedical article published and subsequently recorded in the PubMed database to facilitate retrieving relevant information. With the rapid growth of the PubMed database, large-scale biomedical document indexing becomes increasingly important. MeSH indexing is a challenging task for machine learning, as it needs to assign multiple labels to each article from an extremely large hierachically organized collection. To address this challenge, 
we propose KenMeSH, an end-to-end model that combines new text features and a dynamic \textbf{K}nowledge-\textbf{en}hanced mask attention that integrates document features with MeSH label hierarchy and journal correlation features to index MeSH terms. Experimental results show the proposed method 
achieves state-of-the-art performance on a number of measures.   
\end{abstract}

\section{Introduction}
The PubMed\footnote{\urlstyle{same}\url{https://pubmed.ncbi.nlm.nih.gov/about/}} database is a resource that provides access to the MEDLINE bibliographic database of references and abstracts together with the full text articles of some of these citations which are available in the PubMed Central\footnote{\urlstyle{same}\url{https://en.wikipedia.org/wiki/PubMed\_Central}} (PMC) repository.
MEDLINE\footnote{\urlstyle{same}\url{https://www.nlm.nih.gov/medline/medline_overview.html}} contains more than 28 million references (as of Feb.~2021) to journal articles in the biomedical, health, and related disciplines. Journal articles in MEDLINE are indexed according to \textbf{Me}dical \textbf{S}ubject \textbf{H}eadings (MeSH)\footnote{\urlstyle{same}\url{https://www.nlm.nih.gov/mesh/meshhome.html}}, an hierarchically organized vocabulary that has been developed and maintained by the National Library of Medicine (NLM)\footnote{\urlstyle{same}\url{https://www.nlm.nih.gov}}. Currently, there are 29,369 main MeSH headings, and each MEDLINE citation has 13 MeSH indices, on average. MeSH terms are distinctive features of MEDLINE and can be used in many applications in biomedical text mining and information retrieval \cite{Lu2008EvaluationOQ, Huang2011RecommendingMT, 6374265}, being recognized as important tools for research (e.g., knowledge discovery and hypothesis generation). 

Currently, MeSH indexing is done by human annotators who examine full articles and assign MeSH terms to each article according to rules set by NLM\footnote{\urlstyle{same}\url{https://www.nlm.nih.gov/bsd/indexing/training/TIP_010.html}}. Human annotation is time consuming and costly -- the average cost of annotating one article in MEDLINE is about \$9.40 \cite{Mork2013TheNM}. 
Nearly 1 million citations were added to MEDLINE in 2020 (approximately 2,600 on a daily basis)\footnote{\urlstyle{same}\url{https://www.nlm.nih.gov/bsd/medline_pubmed_production_stats.html}}. The rate of articles being added to the MEDLINE database is constantly increasing, so there is a huge financial and time-consuming cost for the {\em status quo}. Therefore, it is imperative to develop an automatic annotation system that can assist MeSH indexing of large-scale biomedical articles efficiently and accurately.  

Automatic MeSH indexing can be regarded as an extreme multi-label text classification (XMC) problem, where each article can be labeled with multiple MeSH terms. Compared with standard multi-label problems, XMC finds relevant labels from an enormous set of candidate labels. The challenge of large-scale MeSH indexing comes from both the label and article sides. Currently, there are more than 29,000 distinct MeSH terms, and new MeSH terms are updated to the vocabulary every year. The frequency of different MeSH terms appearing in documents are quite imbalanced. For instance, the most frequent MeSH term, `humans', appears in more than 8 million citations; `Pandanaceae', on the other hand,  appears in only 31 documents \cite{10.1093/bioinformatics/btv237}. In addition, the MeSH terms that have been assigned to each article varies greatly, ranging from more than 30 to fewer than 5. Furthermore, semantic features of the biomedical literature are complicated to capture, as they contain many domain-specific concepts, phrases, and abbreviations. The aforementioned  difficulties make the task more complicated to generate an effective and efficient prediction model for MeSH indexing. 

In this work, inspired by the rapid development of deep learning, we propose a novel neural architecture called KenMeSH (\textbf{K}nowledge-\textbf{en}hanced MeSH labelling) which is suitable for handling XMC problems where the labels are arrayed hierarchically and could capture useful information as a directed graph. Our method uses a dynamic knowledge-enhanced mask attention mechanism and incorporates document features together with label features to index biomedical articles. Our major contributions are: 
\begin{enumerate}[noitemsep]
    \item We design a multi-channel document representation module to extract document features from the title and the abstract using a bidirectional LSTM. We use multi-level dilated convolution to capture semantic units in the abstract channel. This module combines a hybrid of information, at the levels of words and the latent representations of the semantic units, to capture local correlations and long-term dependencies from text.  
    \item Our proposed method appears to be the first to employ graph convolutional neural networks that integrate information from the complete MeSH hierarchy to map label representations. 
    \item We propose a novel dynamic knowledge-enhanced mask attention mechanism which incorporates external journal-MeSH co-occurrence information and document similarity in the PubMed database to constrain the large universe of possible labels in the MeSH indexing task.
    \item We evaluate our model on a corpus of PMC articles. Our proposed method consistently achieves superior performance over previous approaches on a number of measures.
\end{enumerate}

\section{Related Work}
\subsection{Automatic MeSH Indexing}
To address the MeSH indexing task mentioned in above section, the National Library of Medicine developed Medical Text Indexer (MTI) -- software that automatically recommends MeSH terms to each MEDLINE article using the abstract and title as input \cite{Aronson2004TheNI}. It first generates the candidate MeSH terms for given articles, and then ranks the candidates to provide the final predictions. There are two modules in MTI -- MetaMap Indexing (MMI) and PubMed-Related Citations (PRC) \cite{Lin2007, 10.1136/jamia.2009.002733}. MetaMap is NLM-developed software which extracts the biomedical concepts in the documents and maps them to Unified Medical Language System concepts. MMI recommends MeSH terms using the biomedical concepts discovered by MetaMap. PRC uses $k$-nearest neighbours to find the MeSH annotations of similar citations in MEDLINE. The two mentioned sets of MeSH terms combine the final MeSH recommendations from MTI.  

BioASQ\footnote{\urlstyle{same}\url{http://bioasq.org}}, an EU-funded project, has organized challenges on automatic MeSH indexing since 2013, which provides opportunities to involve more participants in continuing to the development of MeSH indexing systems. Many effective MeSH indexing systems have been developed since then, such as MeSHLabeler \cite{Liu2015MeSHLabelerIT}, DeepMeSH \cite{Peng2016DeepMeSHDS}, AttentionMeSH \cite{Indexer2018AttentionMeSHS}, and MeSHProbeNet \cite{Xun2019MeSHProbeNetAS}. MeSHLabeler introduced a Learning-to-Rank (LTR) framework, which is a two-step strategy, first predicting the candidate MeSH terms and then ranking them to obtain the final suggestions. MeSHLabeler first trained an independent binary classifier for each MeSH term and then used various evidence, including similar publications and term frequencies, to rank candidate MeSH terms. DeepMeSH is an improved version of MeSHLabeler, which also uses the LTR strategy. It first generates MeSH predictions by incorporating deep semantics in the word embedding space, and then ranks the candidates. AttentionMeSH and MeSHProbeNet are based on bidirectional recurrent neural networks (RNNs) and attention mechanisms. The main difference between AttentionMeSH and MeSHProbeNet is that the former uses a label-wise attention mechanism while the latter develops self-attentive MeSH probes to extract comprehensive aspects of information from the input articles. 

Studies in MeSH indexing with full texts are very limited because of restrictions on full text access. \citet{JimenoYepes2013ComparisonAC} randomly selected 1413 articles from the PMC Open Access Subset and used automatically-generated summaries from these full texts as input to MTI for MeSH indexing. \citet{DemnerFushman2015ExtractingCO} collected 14,828 full text articles from PMC Open Access Subset and developed a rule-based string-matching algorithm to extract a subject of MeSH terms called `check tags' that are used to describe the characteristics of the subjects. \citet{Wang2019IncorporatingFC} randomly selected 257,590 full text articles from PMC Open Access Subset and developed a multi-channel model using CNN-based feature selection to extract important information from different sections of the articles. HGCN4MeSH \cite{Yu2020HGCN4MeSHHG} used the PMC dataset generated by \citet{Wang2019IncorporatingFC} and employed graph convolutional neural network to learn the co-occurrences between MeSH terms. FullMeSH \cite{10.1093/bioinformatics/btz756} and BERTMeSH \cite{You2020BERTMeSHDC} used all available full text articles in PMC Open Access Subset. FullMeSH applied an attention-based CNN to predict the MeSH terms and LTR to get the final MeSH candidates; BERTMeSH incorporated pre-trained BERT and an attention mechanism to improve the performance of MeSH indexing. 

\subsection{Graph Convolutional Networks in Natural Language Processing}
Graph convolutional neural networks (GCN)s \cite{Kipf2017SemiSupervisedCW} have received considerable attention and achieved remarkable success in natural language processing recently. 

Some text classification systems introduce GCN by formulating their problems as graph-structural tasks. For instance, TextGCN \cite{Yao2019GraphCN} built a single text graph for a corpus based on word co-occurrence and document word relations to infer labels. \citet{zhang-etal-2019-aspect} built a GCN-based dependency tree of a sentence to exploit syntactical information and word dependencies for sentiment analysis. Other research focused on learning the relationships between nodes in a graph, such as the label co-occurrences for multi-label text classifications; e.g., MAGNET \cite{Pal2020MultiLabelTC} built a label graph to capture dependency structures among labels, and \citet{rios-kavuluru-2018-shot} built a multi-label classifier that was learned from a 2-layer GCN over the label hierarchy.  

GCN also provides a powerful toolkit for embedding the taxonomies into low dimension representations that could be utilized for specific tasks. For instance, \citet{pujary2020disease} used GCN to learn an undirected graph derived from disease names in the MeSH taxonomy in order to detect and normalize disease mentions in biomedical texts.

\begin{figure*}
\begin{center}
\includegraphics[width=\textwidth, height=9.3cm]{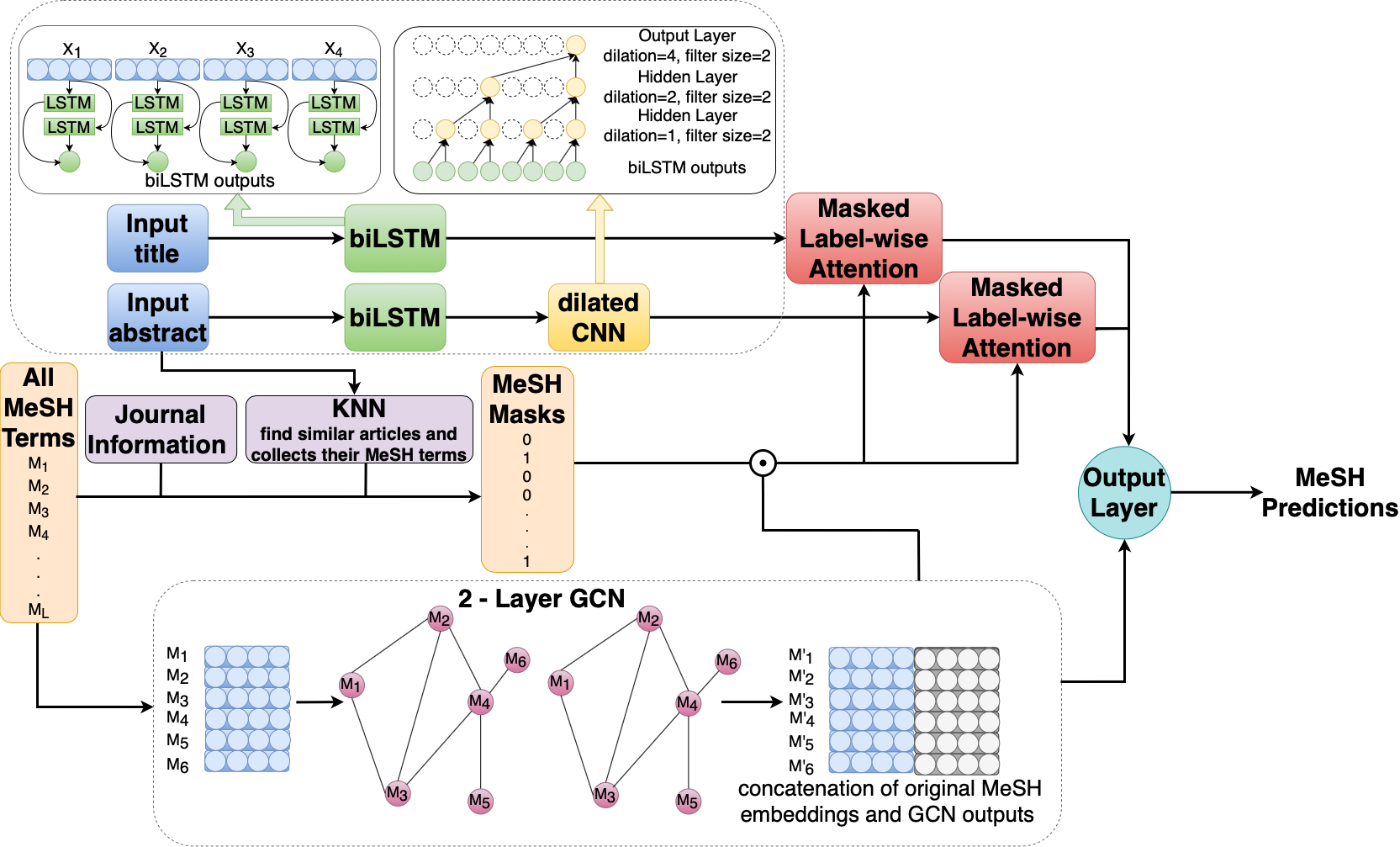}
\caption{Model Architecture - There are three main components in our method. First, a multi-channel document representation module operates on the title and abstract of an input article. Second, a 2-layer GCN creates label vectors. Lastly, a masked attention component calculates the label-specific attention vectors used for predictions.}\label{fig:1}
\end{center}
\end{figure*} 
\section{Proposed Model}
MeSH indexing can be regarded as a multi-label text classification problem in which, given a set of biomedical documents $\mathcal{X}= \{x_{1}, x_{2}, ..., x_{n}\}$ and a set of MeSH labels $\mathcal{Y} = \{y_{1}, y_{2}, ..., y_{L}\}$, multi-label classification learns the function $f: \mathcal{X}  \rightarrow [0, 1]^{\mathcal{Y}}$ using the training set $\mathcal{D} = {(x_{i}, Y_{i})}$, $i = 1, ..., n$, where $n$ is the number of documents in the set. 

Figure \ref{fig:1} illustrates our overall architecture. Our model is composed of a multi-channel document representation module, a label features learning module, a dynamic semantic mask attention module, and a classifier. 

\subsection{Multi-channel Document Representation Module} 
The multi-channel document representation module has two input channels -- the title channel and the abstract channel, for each type of text. These two texts are represented by two embedding matrices, namely $E_{\textit{title}} \in \mathbb{R}^{d}$, the word embedding matrix for the title, and $E_{\textit{abstract}}\in \mathbb{R}^{d}$, the word embedding matrix for the abstract. We first apply a bidirectional Long Short-Term Memory (biLSTM) network \cite{Hochreiter1997LongSM} in both channels to encode the two types of text and to generate the hidden representations $h_{t}$ for each word at time step $t$. The computations of $\overrightarrow{h_{t}}$ and $\overleftarrow{h_{t}}$ are illustrated below:
\begin{equation} \label{eq:1}
    \begin{aligned}
        \overrightarrow{h_{t}} = LSTM(x_{t}, \overrightarrow{h_{t-1}}, c_{t-1}) \\
        \overleftarrow{h_{t}} = LSTM(x_{t}, \overleftarrow{h_{t-1}}, c_{t-1})
    \end{aligned}
\end{equation}
We then obtain the final representation for each word by concatenating the hidden states from both directions, namely $h_{t} = [\overrightarrow{h_{t}}: \overleftarrow{h_{t}}]$ and $h_{t} \in \mathbb{R}^{l \times 2d_{h}}$, where $l$ is the number of words in the text and $d_{h}$ is the hidden dimensions. The biLSTM returns context-aware representations $H_{\textit{title}}$ and $H_{\textit{abstract}}$ for the title and abstract channels, respectively: 
\begin{equation} \label{eq:2}
    \begin{gathered}
        H_{\textit{title}} = biLSTM(E_{\textit{title}}) \\
        H_{\textit{abstract}} = biLSTM(E_{\textit{abstract}})
    \end{gathered}
\end{equation}
In order to generate high-level semantic representations of abstracts, we introduce a dilated convolutional neural network (DCNN) to the abstract channel. The concept of dilated convolution was originally developed for wavelet decomposition \cite{10.1007/978-3-642-75988-8_28}, and has been applied to NLP tasks such as neural machine translation \cite{kalchbrenner2017neural} and text classification \cite{lin-etal-2018-semantic-unit}. The main idea of DCNN is to insert `holes' in convolutional kernels, which  extract the longer-term dependencies and generate higher-level representations, such as phases and sentences. Following \citet{lin-etal-2018-semantic-unit}, we apply a multi-level DCNN with different dilation rates on top of the hidden representations generated by the biLSTM on the abstract channel. Small dilation rates capture phrase-level  information, and large ones capture sentence-level information. The DCNN returns the semantic features of the abstract channel $D_{\textit{abstract}} \in \mathbb{R}^{(l - s + 1) \times 2d_{h}}$, where $s$ is the width of the convolution kernels.

\subsection{Label Features Learning Module}
MeSH taxonomies are organized in 16 categories, and each is further divided into subcategories. Within each subcategory, MeSH terms are ordered hierarchically from most general to most specific, up to 13 hierarchical levels. As the MeSH hierarchy is important to our task, we use a two-layer GCN to incorporate the hierarchical parent and child information among labels. 
We first use the MeSH descriptors to generate a label feature vector for each MeSH term. Each label vector is calculated by averaging the word embedding of each word in its descriptors:
\begin{equation} \label{eq:3}
    v_{i} = \frac{1}{N}\sum_{j \in N}w_{j}, i=1, 2, ..., L,
\end{equation}
where $v_{i} \in \mathbb{R}^{d}$, $N$ is the number of words in its descriptor, and $L$ is the number of labels. In the graph structure, we formulate each node as a MeSH label, and edges represent relationships in the MeSH hierarchy. The edge types of a node include edges from its parent, from its children, and from itself. At each GCN layer, the node feature is aggregated by its parent and children to form the new label feature for the next layer:
\begin{equation} \label{eq:4}
    h^{l+1} = \sigma(A \cdot h^{l} \cdot W^{l}),
\end{equation}
where $h^{l}$ and $h^{l+1} \in \mathbb{R}^{L \times d}$ indicate the node presentation of the $l^{th}$ and $(l+1)^{th}$ layers, $\sigma(\cdot)$ denotes an activation function, $A$ is the adjacency matrix of the MeSH hierarchical graph, and $W^{l}$ is a layer-specific trainable weight matrix. We then concatenate the label feature vectors from descriptors in Equation \ref{eq:3} with GCN label vectors to form:
\begin{equation} \label{eq:5}
    H_{\textit{label}} = [v : h^{l+1}],
\end{equation}
where $H_{\textit{label}} \in \mathbb{R}^{L \times 2d}$ is the final label vector. 

\subsection{Dynamic Knowledge-enhanced Mask Attention Module}
In the dynamic knowledge-enhanced mask attention module, we integrate external knowledge from outside sources to generate a unique mask for each article dynamically. We consider only a subset of the full MeSH list by employing a masked label-wise attention that computes the element-wise multiplication of a mask matrix and an attention matrix for two reasons. First, the MeSH terms are numerous and  have widely varying occurrence frequencies. Therefore, for each MeSH label, there are far more negative examples than positive ones. For each article, selecting a subset of MeSH labels, namely a MeSH mask, down-samples the negative examples, which forces the classifier to concentrate on the candidate labels. Second, the issue with the original attention mechanism \cite{Bahdanau2015NeuralMT} is that the classifier focuses on spotting relevant information for all predicted labels, which is a lack of pertinence. Using a masked label-wise attention allows the classifier to find relevant information for each label inside the MeSH mask.

The dynamic ensures that the module generates a unique MeSH mask for each article, specifically. To generate the MeSH masks, we consider two external knowledge sources: journal information and document similarity. The journal information refers to the name of the journal in which an article was published, which usually defines a specific research domain. We expect that articles published in the same journal tend to be indexed with MeSH terms that are relevant to the journal's research focus. We build a journal--MeSH label co-occurrence matrix using conditional probabilities, i.e., $P(L_{i}\,|\,J_{j})$, which denote the probabilities of occurrence of label $L_{i}$ when journal $J_{j}$ appears. 
\begin{equation} \label{eq:6}
    P(L_{i}\,|\,J_{j}) = \frac{C_{L_{i}\cap J_{j}}}{C_{J_{j}}},
\end{equation}
where $C_{L_{i}\cap J_{j}}$ denotes the number of co-occurrences of $L_{i}$ and $J_{j}$, and $C_{J_{j}}$ is the number of occurrences of $J_{j}$ in the training set. To avoid the noise of rare co-occurrences, a threshold $\tau$ filters noisy correlations.
$M_{j}$ denotes the MeSH label set for journal $j$. 
\begin{equation} \label{eq:7}
    M_{j} = \{L_{k} \vert P(L_{k} \vert J_{j}) > \tau, \;
    k = 1, ..., L\}
\end{equation}
We then use $k$-nearest neighbors (KNN) to choose a subset of specific MeSH terms for each article by referring to document similarity. We represent each article by the IDF-weighted sum of word embeddings in the abstract:
\begin{equation} \label{eq:8}
    D_{\textit{idf}} = \frac{\sum_{i=1}^{n}\textit{IDF}_{i} \times e_{i}}{\sum_{i=1}^{n}\textit{IDF}_{i}},
\end{equation}
where $e_{i}$ is the word embedding, and $\textit{IDF}_{i}$ is the inverse document frequency of the word. Next, 
we use KNN based on cosine similarity between abstracts to find the $K$ nearest neighbours for each article in the training set. To form the unique MeSH mask for article $a$, we collect MeSH terms $M_{a}$ from the neighbours of $a$:
\begin{equation} \label{eq:9}
    M_{a} = T_{1} \cup T_{2} \cup ... \cup T_{K}, 
\end{equation}
where $T_{i}$ is the MeSH label set from the $i^{th}$ neighbour of article $a$. We then join the MeSH labels generated from journal--MeSH co-occurrence for the journal that article $a$ has been published in together with the MeSH terms obtained from the neighbours of article $a$ to form the final MeSH mask label set $M$:
\begin{equation} \label{eq:10}
    M = M_{j} \cup M_{a}
\end{equation}
Then we assign a value to each label in $\mathcal{Y}$ to form $M_{\textit{vec}} \in [0,1]^{\mathcal{Y}}$. If the label appears in $M$, we assign 1, 0 otherwise. The label order of $M_{\textit{vec}}$ is the same as $H_{\textit{label}}$.

We calculate the similarity between MeSH terms and the texts in two channels by applying masked label-wise attention. 
\begin{equation} \label{eq:11}
    \begin{gathered}
        H_{\textit{masked}} = H_{\textit{label}} \odot M_{\textit{vec}} \\
        \alpha_{\textit{title}} = \textrm{Softmax}(H_{\textit{title}} \cdot H_{\textit{masked}}) \\
        \alpha_{\textit{abstract}} = \textrm{Softmax}(D_{\textit{abstract}} \cdot H_{\textit{masked}}),
    \end{gathered}
\end{equation}
where $\odot$ denotes element-wise multiplication, $H_{\textit{masked}}$ denotes the masked label features, and $\alpha_{\textit{title}}$ and $\alpha_{\textit{abstract}}$ measure how informative each text fragment is for each label in the title and  abstract channels, respectively. We then generate the label-specific title and abstract representations, respectively:
\begin{equation} \label{eq:12}
    \begin{gathered}
        c_{\textit{title}} = \alpha_{\textit{title}}^{T} \cdot H_{\textit{title}} \\
        c_{\textit{abstract}} = \alpha_{\textit{abstract}}^{T} \cdot D_{\textit{abstract}},
    \end{gathered}
\end{equation}
such that $c_{\textit{title}} \in \mathbb{R}^{L \times 2d}$, and $c_{\textit{abstract}} \in \mathbb{R}^{L \times 2d}$. We sum up the representations in the title  and abstract channels to form the document vector for each article:
\begin{equation} \label{eq:13}
    D = c_{\textit{title}} + c_{\textit{abstract}}
\end{equation}
\subsection{Classifier}
We gain scores for each MeSH term $i$:
\begin{equation} \label{eq:14}
    \hat{y_{i}} = \sigma(D \odot H_{\textit{label}}), i = 1, 2, ..., L,
\end{equation}
where $\sigma(\cdot)$ represents the sigmoid function. We train our model using the multi-label binary cross-entropy loss \cite{Nam2014LargeScaleMT}:
\begin{equation} \label{eq:15}
    L =\sum_{i=1}^{L}[-y_{i} \cdot log(\hat{y_{i}}) -  (1-y_{i}) \cdot log(1 - \hat{y_{i}}))],
\end{equation}
where $y_{i} \in [0,1]$ is the ground truth of label $i$, and $\hat{y_{i}} \in [0,1]$ denotes the prediction of label $i$ obtained from the proposed model.

\begin{table}[t]
\centering
\resizebox{\columnwidth}{!}{
\begin{tabular}{| c | c | c | c | c | c | c |}
\hline
\textit{Method} & \multicolumn{3}{c|}{\textit{Micro-average Measure}} &\multicolumn{3}{c|}{\textit{Example Based Measure}}\\
\hline
& \textit{MiF} & \textit{MiP} & \textit{MiR} & \textit{EBF} & \textit{EBP} & \textit{EBR}\\
\hline
\textit{MTI} & 0.390 & 0.379 & 0.402 & 0.393 & 0.378 & 0.408\\
\hline
\textit{HGCN4MeSH} & 0.524 & 0.763 & 0.399 & 0.529 & 0.762 & 0.405\\
\hline
\textit{DeepMeSH} & 0.639 & 0.669 & 0.612 & 0.631 & 0.667 &0.627 \\
\hline
\textit{BERTMeSH} & 0.667 & 0.696 & 0.640 & 0.657 & 0.700 & 0.650\\
\hline
\textit{FullMeSH (Full)} & 0.651 & 0.683 & 0.623 & 0.643 & 0.680 & 0.639\\
\hline
\textit{BERTMeSH (Full)} & 0.685 & 0.713 & \textbf{0.659} & 0.675 & 0.717 & \textbf{0.667}\\
\hline
\multirow{2}{*}{\centering\textit{KenMeSH}} & \textbf{0.745} & \textbf{0.864} & 0.655 & \textbf{0.738} & \textbf{0.863} & 0.644\\
~ & $\pm$0.021 &  $\pm$0.011 & $\pm$0.027 & $\pm$0.018 & $\pm$0.011 & $\pm$0.022 \\
\hline
\end{tabular}}
\caption{Comparison to previous methods across two main evaluation metrics. Methods marked as \textit{Full} are trained on entire PMC articles, others on abstracts and titles only. Bold: best scores in each column.}\label{table:1}
\end{table}

\begin{table}[t]
\centering
\resizebox{\columnwidth}{!}{
\begin{tabular}{| c  c | c | c |}
\hline
\multicolumn{2}{|c|}{\textit{Ranking Based}} & 
\multicolumn{2}{|c|}{\multirow{2}{*}{\textit{Methods}}} \\ 

\multicolumn{2}{|c|}{\textit{Measure}}  & \multicolumn{2}{|c|}{} \\
\hline
&  & \textit{HGCN4MeSH} & \textit{KenMeSH}\\
\hline
\multirow{5}{*}{\centering \textit{P@k}}&\textit{$P@1$}& 0.961 & \textbf{0.993$\pm$0.001} \\
~ & \textit{$P@3$} & 0.870 & \textbf{0.972$\pm$0.005} \\
~ & \textit{$P@5$} & 0.788 & \textbf{0.937$\pm$0.010} \\
~ & \textit{$P@10$} & 0.620 & \textbf{0.801$\pm$0.015} \\
~ & \textit{$P@15$} & 0.501 & \textbf{0.659$\pm$0.013} \\\hline
\multirow{5}{*}{\centering \textit{R@k}}&{$R@1$}& 0.077 & \textbf{0.081$\pm$0.000} \\
~ & \textit{$R@3$} & 0.204 & \textbf{0.234$\pm$0.001} \\
~ & \textit{$R@5$} & 0.302 & \textbf{0.370$\pm$0.005} \\
~ & \textit{$R@10$} & 0.460 & \textbf{0.603$\pm$0.012} \\
~ & \textit{$R@15$} & 0.549 & \textbf{0.722$\pm$0.014} \\\hline
\end{tabular} }
\caption{Comparison to HGCN4MeSH across ranking based measures. Bold: best scores in each row.}\label{table:2}
\end{table}
\section{Experiment}

\subsection{Datasets}
We follow \citet{10.1093/bioinformatics/btz756} and \citet{You2020BERTMeSHDC} by using the PMC FTP service\footnote{\urlstyle{same}\url{https://www.ncbi.nlm.nih.gov/research/ bionlp/APIs/BioC-PMC}} \cite{Comeau2019PMCTM} and downloading PMC Open Access Subset (as of Sep.~2021), totalling 3,601,092 citations. We also download the entire MEDLINE collection based on the PubMed Annual Baseline Repository (as of Dec.~2020) and obtain 31,850,051 citations with titles and abstracts. In order to reduce bias, we only focus on articles that are annotated by  human curators (not annotated by a `curated' or `auto' modes in MEDLINE). We then match PMC articles with the citations in PubMed to PMID and obtain a set of 1,284,308 citations. Out of  these PMC articles, we use the latest 20,000 articles as the test set, the next latest 200,000 articles as the validation data set, and the remaining 1.24M articles as the training set. In total, 28,415 distinct MeSH terms are covered in the training dataset.

\subsection{Implementation Details}
We implement our model in PyTorch \cite{NEURIPS2019_9015}. For pre-processing, we removed non-alphanumeric characters, stop words, punctuation, and single character words, and we converted all words to lowercase. Titles longer than 100 characters and abstracts longer than 400 characters are truncated. We use pre-trained biomedical word embeddings (BioWordVec) \cite{Zhang2019BioWordVecIB}, and the embedding dimension is 200. To avoid overfitting, we use dropout directly after the embedding layer with a rate of $0.2$. The number of units in hidden layers are 200 in all three modules. We use a three-level dilated convolution with dilation rate $[1, 2, 3]$ and select 1000 nearest documents to generate MeSH masks for each article. We use FAISS \cite{johnson2019billion} to find  similar documents for each citation among the training set, and the whole process takes 10 hours. We use Adam optimizer \cite{Kingma2015AdamAM} and early stopping strategies. The learning rate is initialized to $0.0003$, and the decay rate is $0.9$ in every epoch. The gradient clip is applied to the maximum norm of 5. The batch size is 32. The model trained for 50 hours on a single NVIDIA V100 GPU. The detailed hyper-parameter settings are shown in Table \ref{table:3}. The code for our method is available at \url{https://github.com/xdwang0726/KenMeSH}.

\begin{table}[ht] 
\begin{center}
\resizebox{\columnwidth}{!}{
\begin{tabular}{| c | l |}

      \hline
      \textit{Hyper-parameters} & \textit{Values}\\
      \hline
      \textit{embedding size} & 200\\
      \hline
      \textit{hidden size} & 200\\
      \hline
      \textit{prediction threshold} & 0.0005\\
      \hline
      \textit{dropout} & \textbf{0.2}, 0.5 \\
      \hline
      \textit{dilation rate} & \textbf{[1, 2, 3]}, [2, 5, 9]\\
      \hline
      \textit{learning rate} &  0.001, 0.0001, \textbf{0.0003}, 0.0005\\
      \hline
      \textit{decay rate} & 0.8, \textbf{0.9}\\
      \hline
      \textit{batch size} & 8, 16, \textbf{32}\\
      \hline
\end{tabular}}
\caption{Hyper-parameter settings. Bold: the optimal values.} \label{table:3}
\end{center}
\end{table}

\subsection{Evaluation Metrics}
We use three main evaluation metrics to test the performance of MeSH indexing systems: Micro-average measure (MiM), example-based measure (EBM), and ranking-based measure (RBM), where MiM and EBM are commonly used in MeSH indexing tasks and RBM is commonly used in evaluating multi-label classification. Micro-average F-measure (MiF) aggregate the global contributions of all MeSH labels and then calculate the harmonic mean of micro-average precision (MiP) and micro-average recall (MiR), which are heavily influenced by frequent MeSH terms. Example-based measures are computed per data point, which computes the harmonic mean of standard precision (EBP) and recall (EBR) for each data point. In the ranking-based measure, precision at $k$ ($P@k$) shows the number of relevant MeSH terms that are suggested in the top-$k$ recommendations of the MeSH indexing system, and recall at $k$ ($R@k$) indicates the proportion of relevant items that are suggested in the top-$k$ recommendations. The detailed computations of evaluation metrics can be found in Appendix \ref{sec:appendixa}.

The threshold has a large influence on MiF and EBF, see Appendix \ref{sec:appendixb}. We select final MeSH labels whose predicted probability is larger than a tuned threshold $t_{i}$:
\begin{equation} \label{eq:16}
    \textit{MeSH}_{i} = 
    \begin{cases}
        \hat{y_{i}} \geq t_{i},  1 \\
        \hat{y_{i}} < t_{i}, 0 
    \end{cases} 
\end{equation}
where $t_{i}$ is the threshold for MeSH term $i$. We compute optimal threshold for each MeSH term on the validation set following \citet{PILLAI20132055} that tunes $t_{i}$ by maximizing MiF:
\begin{equation} \label{eq:17}
    t_{i} = \argmax_\textbf{T} \textit{MiF}(\textbf{T}),
\end{equation}
where $\textbf{T}$ denotes all possible threshold values for label $i$.

\begin{figure*}
     \centering
     \begin{subfigure}[b]{0.32\textwidth}
         \centering
         \includegraphics[width=\textwidth, height=4.1cm]{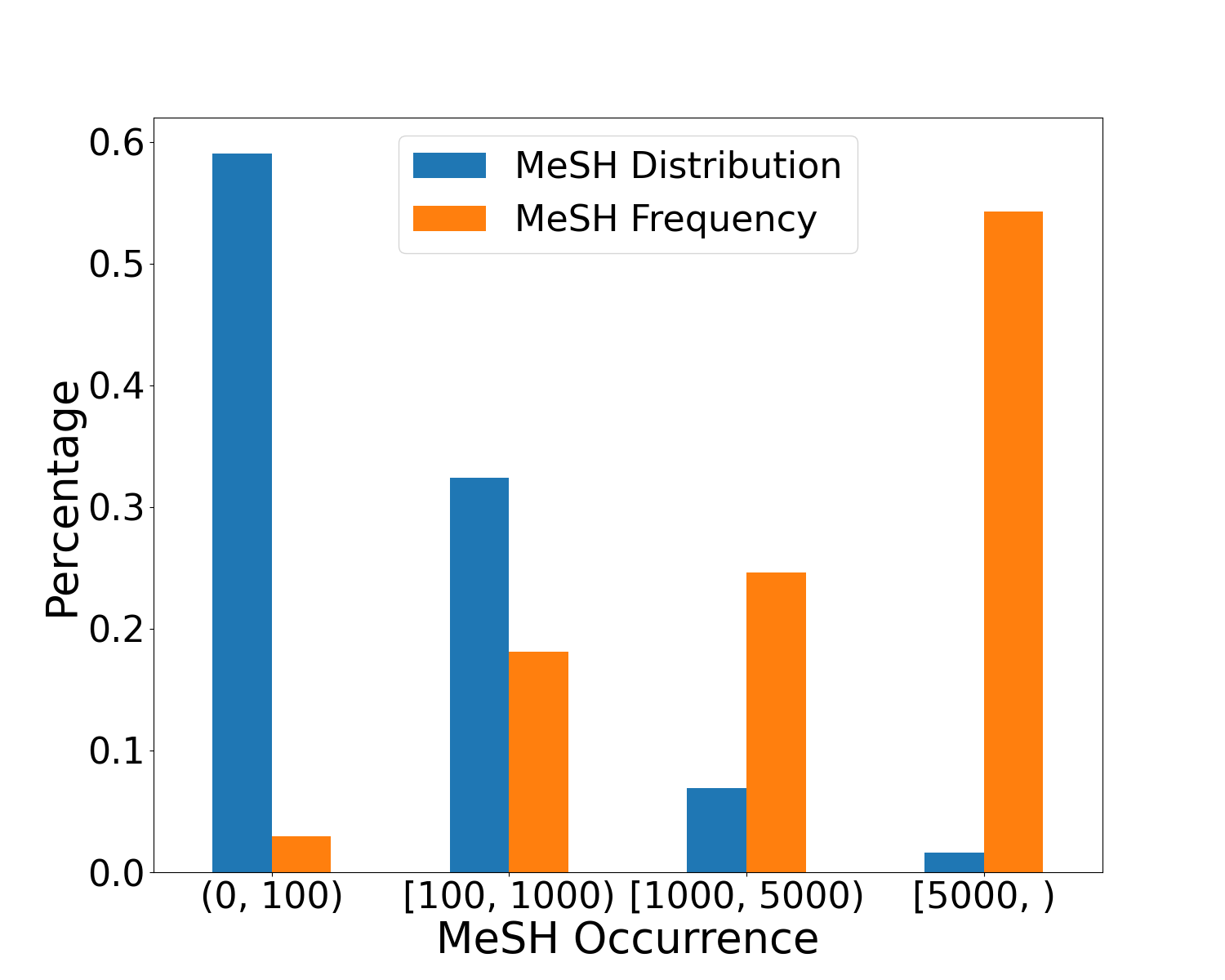}
         \caption{MeSH Terms Distribution}
         \label{fig:mesh_dist}
     \end{subfigure}
     \hfill
     \begin{subfigure}[b]{0.32\textwidth}
         \centering
         \includegraphics[width=\textwidth, height=4.1cm]{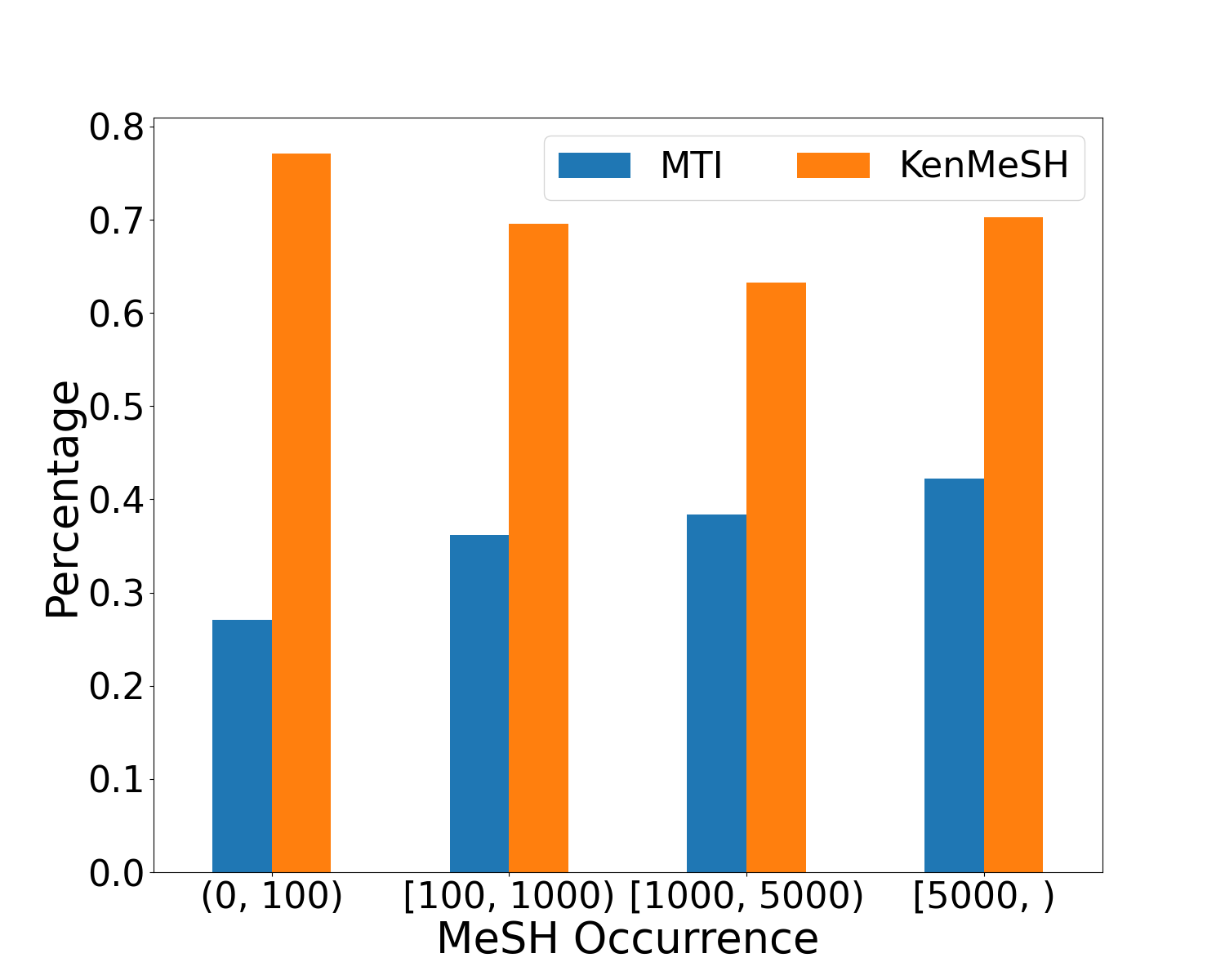}
         \caption{MeSH Performance on MiF}
         \label{fig:infreq_mif}
     \end{subfigure}
     \hfill
     \begin{subfigure}[b]{0.32\textwidth}
         \centering
         \includegraphics[width=\textwidth,  height=4.1cm]{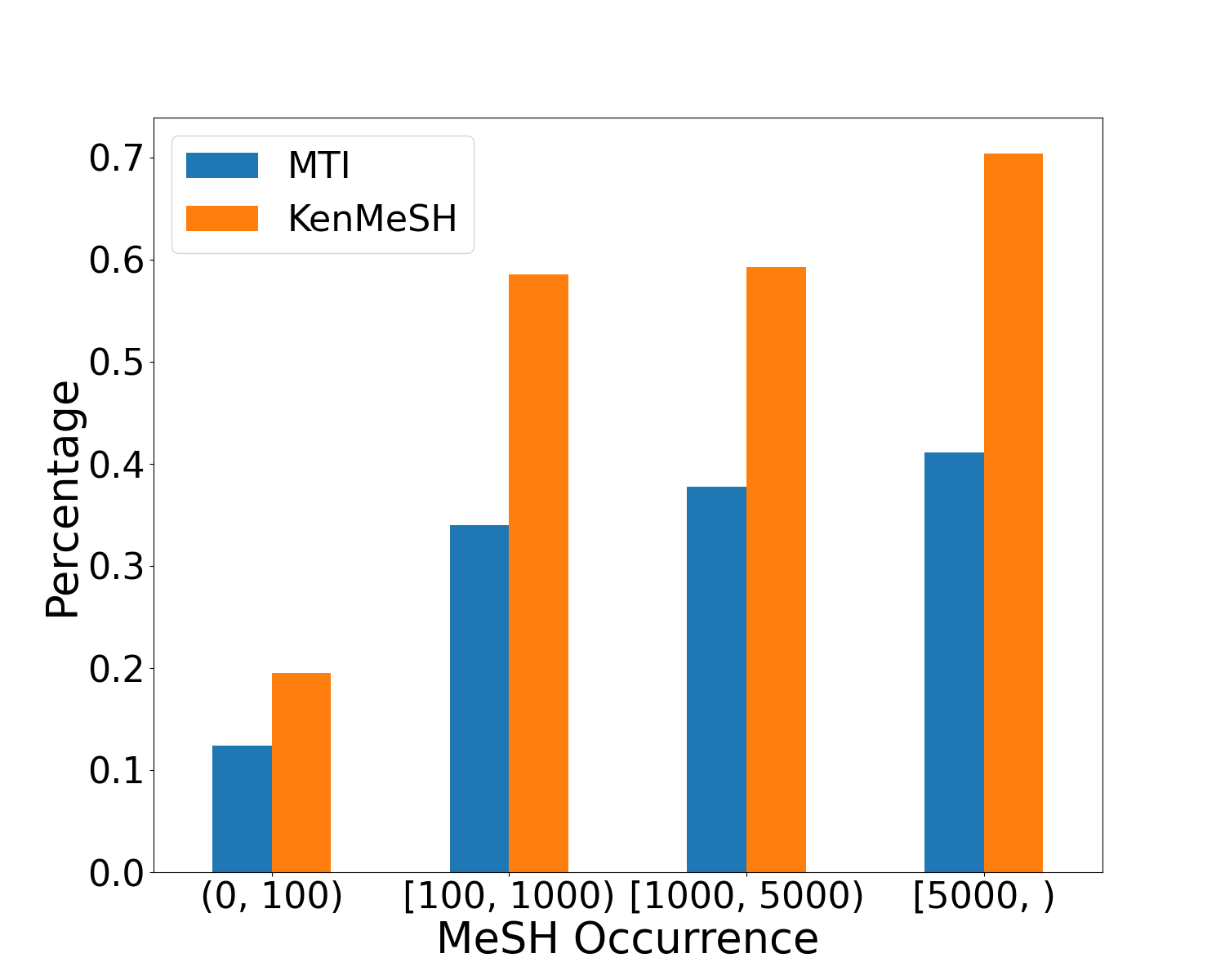}
         \caption{MeSH Performance on EBF}
         \label{fig:infreq_ebf}
     \end{subfigure}
        \caption{Performance comparison of our model and MTI on MeSH terms at different frequency}
        \label{fig:three graphs}
\end{figure*}

\begin{table*}[t]
\centering
\resizebox{\textwidth}{!}{
\begin{tabular}{| c | c | c | c | c | c | c | c | c | c |}
\hline
\textit{Methods} & \multicolumn{3}{c|}{\textit{precision @ k}}& \multicolumn{3}{c|}{\textit{Micro-average Measure}} &\multicolumn{3}{c|}{\textit{Example Based Measure}}\\
\hline
& $p@1$ & $p@3$ & $p@5$ & \textit{MiF} & \textit{MiP} & \textit{MiR} & \textit{EBF} & \textit{EBP} & \textit{EBR}\\
\hline
\textit{Full Model} & \textbf{0.993} & \textbf{0.972} & \textbf{0.936} & \textbf{0.745} & \textbf{0.864} & \textbf{0.655} & \textbf{0.738} & \textbf{0.863} & \textbf{0.644} \\
\hline
\textit{Ablation-(a)} & 0.983 & 0.938 & 0.882 & 0.672 & 0.752 & 0.609 & 0.680 & 0.751 & 0.621 \\
\hline
\textit{Ablation-(b)} & 0.988 & 0.952 & 0.900 & 0.687 & 0.788 & 0.551 & 0.695 & 0.788 & 0.622 \\
\hline
\textit{Ablation-(c)} & 0.968 & 0.893 & 0.816 & 0.554 & 0.789 & 0.427 & 0.548 & 0.791 & 0.419 \\
\hline
\textit{Ablation-(d)} & 0.987 & 0.949 & 0.896 & 0.674 & 0.806 & 0.579 & 0.681 & 0.805 & 0.591 \\
\hline
\end{tabular}}
\caption{Ablation experiment results. (a) Without multi-channel settings, texts and abstracts are in the same channel. (b) Without DCNN on the abstract channel. (c) Without label feature module. (d) Without semantic mask attention module. Bold: best scores.}\label{table:4}
\end{table*}

\section{Results and Ablation Studies}
We evaluate our proposed model with five state-of-the-art models: MTI, DeepMeSH, FullMeSH, BERTMeSH and HGCN4MeSH. Among these, MTI, DeepMeSH, BERTMeSH, and HGCN4MeSH are trained with abstracts and titles only; FullMeSH (Full) and BERTMeSH (Full) are trained with full PMC articles. Our proposed model is trained on titles and abstracts, and is tested using 20,000 of the latest articles. We mainly focus on MiF, which is the main evaluation metric in MeSH indexing task. 

We compare our model against previous related systems on micro-average measure and example-bases measure in Table \ref{table:1}. Each row in the table shows all evaluation metrics on a specific method, where the best score for each metric is indicated. As reported, our model achieves the best performance on most evaluation metrics, expect MiR and EBR, on which BERTMeSH (Full) achieves the best performance. This is because that BERTMeSH (Full) is trained on full text articles, which uses much more content information in the articles than ours. Our model outperforms the subset of systems that were trained only on the abstract and the title --  MTI, HGCN4MeSH, DeepMeSH and BERTMeSH in all metrics. Most importantly, there is improvement in precision without a decrease in recall. 
Comparing with systems trained on full articles indicates that our model achieves the best MiF, and is only slightly below BERTMeSH (Full) on MiR (0.4 percentage points). Although our model is trained only on the abstract and title (which may suggest that it captures less complex semantics), it performs very well against more complex systems. Furthermore, we compare the performance of our model with HGCN4MeSH on ranking-based measures that do not require a specific threshold. The results, summarized in Table \ref{table:2}, show that our model always performs better than HGCN4MeSH with up to almost 18\% improvement. 

As the frequency of different MeSH terms are imbalanced, we are interested in examining the efficiency of our model on infrequent MeSH terms. We divide MeSH terms into four groups based on the number of occurrences in the training set: $(0,100)$, $[100, 1000)$, $[1000, 5000)$, and $[5000, )$. Figure \ref{fig:mesh_dist} shows the distribution of MeSH terms and percent of occurrence among the four divided groups in the training set, which indicates that the distribution of MeSH frequency is highly biased and it falls into a long-tail distribution. Figure \ref{fig:infreq_mif} and \ref{fig:infreq_ebf} show the performance of our model comparing to MTI baseline in the four MeSH groups on MiF and EBF respectively. Our model obtains substantial improvements among frequent and infrequent labels on both MiF and EBF.

We are interested in studying how the effectiveness and robustness of our model are due to the various modules, such as the multi-channel mechanism, the dilated CNN, the label graph, and masked attention. To further understand the impacts of these factors, we conduct controlled experiments with four different settings: (a) examining a single channel architecture by concatenating the title and abstract as input into the abstract channel; (b) removing the dilated CNN; (c) replacing the label feature learning module with a fully connected layer; and (d) removing the masked attention module. The influence of each of these modules can then be evaluated individually. 
The results are summarized in Table \ref{table:4}.
\subparagraph{Impacts on Multi-channel Settings} As shown in Table \ref{table:4}, the multi-channel setting outperforms the single channel one. 
The reason for this could be that the single channel model misses some important features in titles and abstracts in the LSTM layer. LSTM has the capability to learn and remember over long sequences of inputs, but it can be challenging to use when facing very long input sequences. Concatenating the title and abstract into one longer sequence may hurt the performance of LSTM. To be more explicit, the single channel model may be remembering insignificant features in the LSTM layer when dealing with longer sequences. Therefore, extracting information from the title and the abstract separately is better than directly concatenating the information.  
\subparagraph{Impacts on Dilated Semantic Feature Extractions} As reported in Table \ref{table:4}, the performance drops when removing the dilated CNN layer. The reason for this seems to be that multi-level dilated CNNs can extract high-level semantic information from the semantic units that are often wrapped in phrases or sentences, and then capture local correlation together with longer-term dependencies from the text. Compared with word-level information extracted from the biLSTM layer, high-level information extracted from the semantic units seems to provide better understanding of the text, at least for the purposes of labelling. 
\subparagraph{Impacts on Learning Label Features} As shown in Table \ref{table:4}, not learning the label features has the largest negative impacts on performance especially for recall (and subsequently F-measure). By removing the label features, the model pays more attention to the frequent MeSH terms and misclassifies infrequent labels as negative. This indicates that label features learned through GCN can capture the hierarchical information between MeSH terms, and MeSH indexing for infrequent terms can benefit from this hierarchical information.
\subparagraph{Impacts on Dynamic Knowledge-enhanced Mask Attention}Table \ref{table:4} shows a performance drop when removing the masked attention layer, suggesting that the attention mechanism has positive impacts on performance. This result further suggest that the masked attention takes advantage of incorporating external knowledge to alleviate the extremely large pool of possible labels. To select the proper mask for each article, two hyperparameters are used: threshold $\tau$ for journal-MeSH occurrence and the number of nearest articles $K$. With $\tau = 0.5$ and $K=1000$, all of the gold-standard MeSH labels are guaranteed to be in the mask.  

\section{Conclusion}
We propose a novel end-to-end model integrating document features and label hierarchical features for MeSH indexing. We use a novel dynamic knowledge-enhanced mask attention mechanism to handle the large universe of candidate MeSH terms and employ GCN in extracting label correlations. Experimental results demonstrate that our proposed model significantly outperforms the baseline models and provides especially large improvements on infrequent MeSH labels. 

In the future, we believe two important research directions will lead to further improvements. First, we plan to explore full text articles, which contain more information, to see whether our model takes advantage of the full text to improve the performance of large-scale MeSH indexing. Second, we are interested in integrating knowledge from the Unified Medical Language System (UMLS) \cite{Bodenreider2004TheUM}, a comprehensive ontology of biomedical concepts, 
in our model. 

\section*{Acknowledgements}
We thank all reviewers and area chairs for their constructive comments and feedback. Resources used in preparing this research were provided, in part, by Compute Ontario (\url{www.computeontario.ca}), Compute Canada (\url{www.computecanada.ca}), the Province of Ontario, the Government of Canada through CIFAR, and companies sponsoring the Vector Institute (\url{www.vectorinstitute.ai/partners}). This research is partially funded by The Natural Sciences and Engineering Research Council of Canada (NSERC) through a Discovery Grant to R. E. Mercer. F. Rudzicz is supported by a CIFAR Chair in AI.

\bibliography{anthology,custom}

\begin{thebibliography}{37}
\expandafter\ifx\csname natexlab\endcsname\relax\def\natexlab#1{#1}\fi

\bibitem[{Aronson et~al.(2004)Aronson, Mork, Gay, Humphrey, and
  Rogers}]{Aronson2004TheNI}
A.~Aronson, James~G. Mork, Clifford~W. Gay, S.~Humphrey, and Willie~J. Rogers.
  2004.
\newblock {The NLM Indexing Initiative's Medical Text Indexer}.
\newblock \emph{Studies in health technology and informatics}, 107 Pt
  1:268--72.

\bibitem[{Aronson and Lang(2010)}]{10.1136/jamia.2009.002733}
Alan~R Aronson and Fran{\c c}ois-Michel Lang. 2010.
\newblock \href {https://doi.org/10.1136/jamia.2009.002733} {An overview of
  {MetaMap}: Historical perspective and recent advances}.
\newblock \emph{Journal of the American Medical Informatics Association},
  17(3):229--236.

\bibitem[{Bahdanau et~al.(2015)Bahdanau, Cho, and
  Bengio}]{Bahdanau2015NeuralMT}
Dzmitry Bahdanau, Kyunghyun Cho, and Yoshua Bengio. 2015.
\newblock Neural machine translation by jointly learning to align and
  translate.
\newblock \emph{CoRR}, abs/1409.0473.

\bibitem[{Bodenreider(2004)}]{Bodenreider2004TheUM}
Olivier Bodenreider. 2004.
\newblock The unified medical language system (umls): integrating biomedical
  terminology.
\newblock \emph{Nucleic acids research}, 32 Database issue:D267--70.

\bibitem[{Comeau et~al.(2019)Comeau, Wei, Dogan, and Lu}]{Comeau2019PMCTM}
Donald~C. Comeau, Chih-Hsuan Wei, R.~Dogan, and Zhiyong Lu. 2019.
\newblock {PMC text mining subset in BioC: about three million full-text
  articles and growing}.
\newblock \emph{Bioinformatics}.

\bibitem[{Dai et~al.(2019)Dai, You, Lu, Huang, Mamitsuka, and
  Zhu}]{10.1093/bioinformatics/btz756}
Suyang Dai, Ronghui You, Zhiyong Lu, Xiaodi Huang, Hiroshi Mamitsuka, and
  Shanfeng Zhu. 2019.
\newblock \href {https://doi.org/10.1093/bioinformatics/btz756} {{FullMeSH:
  improving large-scale MeSH indexing with full text}}.
\newblock \emph{Bioinformatics}, 36(5):1533--1541.

\bibitem[{Demner-Fushman and Mork(2015)}]{DemnerFushman2015ExtractingCO}
Dina Demner-Fushman and James~G. Mork. 2015.
\newblock Extracting characteristics of the study subjects from full-text
  articles.
\newblock In \emph{Proceedings of the American Medical Informatics Association
  (AMIA) Annual Symposium}, pages 484--491.

\bibitem[{Gu et~al.(2013)Gu, Feng, Zeng, Mamitsuka, and Zhu}]{6374265}
Jun Gu, Wei Feng, Jia Zeng, Hiroshi Mamitsuka, and Shanfeng Zhu. 2013.
\newblock \href {https://doi.org/10.1109/TSMCB.2012.2227998} {{Efficient
  Semisupervised MEDLINE Document Clustering With MeSH-Semantic and
  Global-Content Constraints}}.
\newblock \emph{IEEE Transactions on Cybernetics}, 43(4):1265--1276.

\bibitem[{Hochreiter and Schmidhuber(1997)}]{Hochreiter1997LongSM}
Sepp Hochreiter and J{\"u}rgen Schmidhuber. 1997.
\newblock Long short-term memory.
\newblock \emph{Neural Computation}, 9:1735--1780.

\bibitem[{Holschneider et~al.(1990)Holschneider, Kronland-Martinet, Morlet, and
  Tchamitchian}]{10.1007/978-3-642-75988-8_28}
M.~Holschneider, R.~Kronland-Martinet, J.~Morlet, and Ph. Tchamitchian. 1990.
\newblock A real-time algorithm for signal analysis with the help of the
  wavelet transform.
\newblock In \emph{Wavelets}, pages 286--297, Berlin, Heidelberg. Springer
  Berlin Heidelberg.

\bibitem[{Huang et~al.(2011)Huang, N{\'e}v{\'e}ol, and
  Lu}]{Huang2011RecommendingMT}
Minlie Huang, Aur{\'e}lie N{\'e}v{\'e}ol, and Zhiyong Lu. 2011.
\newblock {Recommending MeSH terms for annotating biomedical articles}.
\newblock \emph{Journal of the American Medical Informatics Association :
  JAMIA}, 18:660 -- 667.

\bibitem[{Jimeno-Yepes et~al.(2013)Jimeno-Yepes, Mork, Demner-Fushman, and
  Aronson}]{JimenoYepes2013ComparisonAC}
Antonio Jimeno-Yepes, James~G. Mork, Dina Demner-Fushman, and Alan~R. Aronson.
  2013.
\newblock Comparison and combination of several {MeSH} indexing approaches.
\newblock In \emph{Proceedings of the American Medical Informatics Association
  (AMIA) Annual Symposium}, pages 709--718.

\bibitem[{Jin et~al.(2018)Jin, Dhingra, and Cohen}]{Indexer2018AttentionMeSHS}
Qiao Jin, Bhuwan Dhingra, and William~W. Cohen. 2018.
\newblock {AttentionMeSH}: Simple, effective and interpretable automatic {MeSH}
  indexer.
\newblock In \emph{Proceedings of the 2018 EMNLP Workshop BioASQ: Large-scale
  Biomedical Semantic Indexing and Question Answering}, pages 47--56.

\bibitem[{Johnson et~al.(2019)Johnson, Douze, and
  J{\'e}gou}]{johnson2019billion}
Jeff Johnson, Matthijs Douze, and Herv{\'e} J{\'e}gou. 2019.
\newblock Billion-scale similarity search with {GPUs}.
\newblock \emph{IEEE Transactions on Big Data}, 7(3):535--547.

\bibitem[{Kalchbrenner et~al.(2017)Kalchbrenner, Espeholt, Simonyan, van~den
  Oord, Graves, and Kavukcuoglu}]{kalchbrenner2017neural}
Nal Kalchbrenner, Lasse Espeholt, Karen Simonyan, Aaron van~den Oord, Alex
  Graves, and Koray Kavukcuoglu. 2017.
\newblock \href {http://arxiv.org/abs/1610.10099} {Neural machine translation
  in linear time}.

\bibitem[{Kingma and Ba(2015)}]{Kingma2015AdamAM}
Diederik~P. Kingma and Jimmy Ba. 2015.
\newblock Adam: A method for stochastic optimization.
\newblock \emph{CoRR}, abs/1412.6980.

\bibitem[{Kipf and Welling(2017)}]{Kipf2017SemiSupervisedCW}
Thomas Kipf and Max Welling. 2017.
\newblock Semi-supervised classification with graph convolutional networks.
\newblock \emph{ArXiv}, abs/1609.02907.

\bibitem[{Lin and Wilbur(2007)}]{Lin2007}
Jimmy Lin and W.~John Wilbur. 2007.
\newblock \href {https://doi.org/10.1186/1471-2105-8-423} {{PubMed related
  articles: A probabilistic topic-based model for content similarity}}.
\newblock \emph{BMC Bioinformatics}, 8(1):423.

\bibitem[{Lin et~al.(2018)Lin, Su, Yang, Ma, and
  Sun}]{lin-etal-2018-semantic-unit}
Junyang Lin, Qi~Su, Pengcheng Yang, Shuming Ma, and Xu~Sun. 2018.
\newblock \href {https://doi.org/10.18653/v1/D18-1485} {Semantic-unit-based
  dilated convolution for multi-label text classification}.
\newblock In \emph{Proceedings of the 2018 Conference on Empirical Methods in
  Natural Language Processing}, pages 4554--4564, Brussels, Belgium.
  Association for Computational Linguistics.

\bibitem[{Liu et~al.(2015)Liu, Peng, Wu, Zhai, Mamitsuka, and
  Zhu}]{Liu2015MeSHLabelerIT}
Ke~Liu, Shengwen Peng, Junqiu Wu, ChengXiang Zhai, Hiroshi Mamitsuka, and
  Shanfeng Zhu. 2015.
\newblock {MeSHLabeler}: Improving the accuracy of large-scale mesh indexing by
  integrating diverse evidence.
\newblock \emph{Bioinformatics}, 31(12):i339--i347.

\bibitem[{Lu et~al.(2008)Lu, Kim, and Wilbur}]{Lu2008EvaluationOQ}
Zhiyong Lu, W.~Kim, and W.~Wilbur. 2008.
\newblock {Evaluation of query expansion using MeSH in PubMed}.
\newblock \emph{Information Retrieval}, 12:69--80.

\bibitem[{Mork et~al.(2013)Mork, Jimeno-Yepes, and Aronson}]{Mork2013TheNM}
James~G. Mork, Antonio Jimeno-Yepes, and Alan~R. Aronson. 2013.
\newblock The {NLM} {M}edical {T}ext {I}ndexer system for indexing biomedical
  literature.
\newblock In \emph{Proceedings of the first Workshop on Bio-Medical Semantic
  Indexing and Question Answering (BioASQ)}.

\bibitem[{Nam et~al.(2014)Nam, Kim, Menc{\'i}a, Gurevych, and
  F{\"u}rnkranz}]{Nam2014LargeScaleMT}
Jinseok Nam, Jungi Kim, Eneldo~Loza Menc{\'i}a, Iryna Gurevych, and Johannes
  F{\"u}rnkranz. 2014.
\newblock Large-scale multi-label text classification - revisiting neural
  networks.
\newblock \emph{ArXiv}, abs/1312.5419.

\bibitem[{Pal et~al.(2020)Pal, Selvakumar, and
  Sankarasubbu}]{Pal2020MultiLabelTC}
Ankit Pal, M.~Selvakumar, and Malaikannan Sankarasubbu. 2020.
\newblock Multi-label text classification using attention-based graph neural
  network.
\newblock In \emph{ICAART}.

\bibitem[{Paszke et~al.(2019)Paszke, Gross, Massa, Lerer, Bradbury, Chanan,
  Killeen, Lin, Gimelshein, Antiga, Desmaison, Kopf, Yang, DeVito, Raison,
  Tejani, Chilamkurthy, Steiner, Fang, Bai, and Chintala}]{NEURIPS2019_9015}
Adam Paszke, Sam Gross, Francisco Massa, Adam Lerer, James Bradbury, Gregory
  Chanan, Trevor Killeen, Zeming Lin, Natalia Gimelshein, Luca Antiga, Alban
  Desmaison, Andreas Kopf, Edward Yang, Zachary DeVito, Martin Raison, Alykhan
  Tejani, Sasank Chilamkurthy, Benoit Steiner, Lu~Fang, Junjie Bai, and Soumith
  Chintala. 2019.
\newblock {PyTorch}: An imperative style, high-performance deep learning
  library.
\newblock In H.~Wallach, H.~Larochelle, A.~Beygelzimer, F.~d\textquotesingle
  Alch\'{e}-Buc, E.~Fox, and R.~Garnett, editors, \emph{Advances in Neural
  Information Processing Systems 32}, pages 8024--8035. Curran Associates, Inc.

\bibitem[{Peng et~al.(2016)Peng, You, Wang, Zhai, Mamitsuka, and
  Zhu}]{Peng2016DeepMeSHDS}
Shengwen Peng, Ronghui You, Hongning Wang, ChengXiang Zhai, Hiroshi Mamitsuka,
  and Shanfeng Zhu. 2016.
\newblock {DeepMeSH: deep semantic representation for improving large-scale
  MeSH indexing}.
\newblock \emph{Bioinformatics}, 32(12):i70--i79.

\bibitem[{Pillai et~al.(2013)Pillai, Fumera, and Roli}]{PILLAI20132055}
Ignazio Pillai, Giorgio Fumera, and Fabio Roli. 2013.
\newblock \href {https://doi.org/https://doi.org/10.1016/j.patcog.2013.01.012}
  {Threshold optimisation for multi-label classifiers}.
\newblock \emph{Pattern Recognition}, 46(7):2055--2065.

\bibitem[{Pujary et~al.(2020)Pujary, Thorne, and Aziz}]{pujary2020disease}
Dhruba Pujary, Camilo Thorne, and Wilker Aziz. 2020.
\newblock \href {http://arxiv.org/abs/2010.12925} {Disease normalization with
  graph embeddings}.

\bibitem[{Rios and Kavuluru(2018)}]{rios-kavuluru-2018-shot}
Anthony Rios and Ramakanth Kavuluru. 2018.
\newblock \href {https://doi.org/10.18653/v1/D18-1352} {Few-shot and zero-shot
  multi-label learning for structured label spaces}.
\newblock In \emph{Proceedings of the 2018 Conference on Empirical Methods in
  Natural Language Processing}, pages 3132--3142, Brussels, Belgium.
  Association for Computational Linguistics.

\bibitem[{Wang and Mercer(2019)}]{Wang2019IncorporatingFC}
Xindi Wang and Robert~E. Mercer. 2019.
\newblock {Incorporating Figure Captions and Descriptive Text in MeSH Term
  Indexing}.
\newblock In \emph{BioNLP@ACL}.

\bibitem[{Xun et~al.(2019)Xun, Jha, Yuan, Wang, and
  Zhang}]{Xun2019MeSHProbeNetAS}
Guangxu Xun, Kishlay Jha, Ye~Yuan, Yaqing Wang, and Aidong Zhang. 2019.
\newblock {MeSHProbeNet: A self-attentive probe net for MeSH indexing}.
\newblock \emph{Bioinformatics}.

\bibitem[{Yao et~al.(2019)Yao, Mao, and Luo}]{Yao2019GraphCN}
Liang Yao, Chengsheng Mao, and Yuan Luo. 2019.
\newblock Graph convolutional networks for text classification.
\newblock In \emph{AAAI}.

\bibitem[{You et~al.(2020)You, Liu, Mamitsuka, and Zhu}]{You2020BERTMeSHDC}
R.~You, Yuxuan Liu, Hiroshi Mamitsuka, and Shanfeng Zhu. 2020.
\newblock Bertmesh: Deep contextual representation learning for large-scale
  high-performance mesh indexing with full text.
\newblock \emph{Bioinformatics}.

\bibitem[{Yu et~al.(2020)Yu, Yang, and Li}]{Yu2020HGCN4MeSHHG}
Miaomiao Yu, Yujiu Yang, and Chenhui Li. 2020.
\newblock {HGCN4MeSH: Hybrid Graph Convolution Network for MeSH Indexing}.
\newblock In \emph{ACL}.

\bibitem[{Zhai et~al.(2015)Zhai, Mamitsuka, Wu, Liu, Zhu, and
  Peng}]{10.1093/bioinformatics/btv237}
Chengxiang Zhai, Hiroshi Mamitsuka, Junqiu Wu, Ke~Liu, Shanfeng Zhu, and
  Shengwen Peng. 2015.
\newblock {MeSHLabeler}: Improving the accuracy of large-scale {MeSH} indexing
  by integrating diverse evidence.
\newblock \emph{Bioinformatics}, 31(12):i339--i347.

\bibitem[{Zhang et~al.(2019{\natexlab{a}})Zhang, Li, and
  Song}]{zhang-etal-2019-aspect}
Chen Zhang, Qiuchi Li, and Dawei Song. 2019{\natexlab{a}}.
\newblock \href {https://doi.org/10.18653/v1/D19-1464} {Aspect-based sentiment
  classification with aspect-specific graph convolutional networks}.
\newblock In \emph{Proceedings of the 2019 Conference on Empirical Methods in
  Natural Language Processing and the 9th International Joint Conference on
  Natural Language Processing (EMNLP-IJCNLP)}, pages 4568--4578, Hong Kong,
  China. Association for Computational Linguistics.

\bibitem[{Zhang et~al.(2019{\natexlab{b}})Zhang, Chen, Yang, Lin, and
  Lu}]{Zhang2019BioWordVecIB}
Yijia Zhang, Qingyu Chen, Zhihao Yang, Hongfei Lin, and Zhiyong Lu.
  2019{\natexlab{b}}.
\newblock {BioWordVec, improving biomedical word embeddings with subword
  information and MeSH}.
\newblock \emph{Scientific Data}, 6.

\end{thebibliography}
\bibliographystyle{acl_natbib}

\appendix

\section{Evaluation Metrics}
\label{sec:appendixa}
Micro F-measure (MiF) computes the harmonic mean of micro-average precision (MiF) and micro-average recall (MiR):
\begin{equation}
    \textit{MiF} =  \frac{2 \times\textit{MiR} \times \textit{MiP}}{\textit{MiR} + \textit{MiP}},
\end{equation}
where 
\begin{equation}
    \textit{MiP} = \frac{\sum_{j=1}^{L}\textit{TP}_{j}}{\sum_{j=1}^{L}\textit{TP}_{j} + \sum_{j=1}^{L}\textit{FP}_{j}},
\end{equation}
\begin{equation}
    \textit{MiR} = \frac{\sum_{j=1}^{L}\textit{TP}_{j}}{\sum_{j=1}^{L}\textit{TP}_{j} + \sum_{j=1}^{L}\textit{FN}_{j}},
\end{equation}
where $\textit{TP}_{j}$, $\textit{FP}_{j}$ and $\textit{FN}_{j}$ as true positives, false positives, and false negatives respectively for each label $l_{j}$ in the set of total labels $L$.

EBF can be computed as the harmonic mean of standard precision (EBP) and recall (EBR):
\begin{equation}
    \textit{EBF} = \frac{2 \times\textit{EBR} \times \textit{EBP}}{\textit{EBR} + \textit{EBP}},
\end{equation}
where
\begin{equation}
    \textit{EBP} = \frac{1}{N}\sum_{i = 1}^{N}\frac{| y_{i} \cap \hat{y_{i}} |}{| \hat{y_{i}} | },
\end{equation}
\begin{equation}
    \textit{EBR}= \frac{1}{N}\sum_{i = 1}^{N}\frac{| y_{i} \cap \hat{y_{i}} |}{| y_{i} | },
\end{equation}
where $y_i$ is the true label set and $ \hat{y_{i}}$ is the predicted label set for instance $i$, $N$ represents the total number of instance. 

Ranking-based evaluation, including precision at \textit{k} (\textit{P@k}), and recall at  \textit{k} (\textit{R@k}). The metrics are defined as follows:
\begin{equation}
    \textit{P@k} = \frac{1}{k}\sum_{l \in r_{k}\left(\hat{y}\right)}y_{l},
\end{equation}
\begin{equation}
    \textit{R@k} = \frac{1}{| y_{i} |}\sum_{l \in r_{k}\left(\hat{y}\right)}y_{l},
\end{equation}
where $r_{k}$ returns the top-$k$ recommended items. 

\section{Threshold Selection Affects the Measurements}
\label{sec:appendixb}
\begin{table}[h]
\centering
\resizebox{\columnwidth}{!}{
\begin{tabular}{| c | c | c | c | c | c | c |}
\hline
\textit{Threshold Values} & \multicolumn{3}{c|}{\textit{Micro-average Measure}} &\multicolumn{3}{c|}{\textit{Example Based Measure}}\\
\hline
& \textit{MiF} & \textit{MiP} & \textit{MiR} & \textit{EBF} & \textit{EBP} & \textit{EBR}\\
\hline
0.5 & 0.707 & 0.908 & 0.579 & 0.716 & 0.907 & 0.592\\
\hline
0.05 & 0.739 & 0.864 & 0.645 & 0.747 & 0.865 & 0.658\\
\hline
0.005 & 0.741 & 0.858 & 0.652 & 0.749 & 0.859 & 0.664 \\
\hline
0.0005 & 0.745 & 0.864 & 0.655 & 0.738 & 0.863 & 0.644\\
\hline
\end{tabular}}
\caption{Comparison to different threshold values across two main evaluation metrics.}\label{table:5}
\end{table}

Thresholds have a huge impact on multi-label evaluation measures. We test the model's performance on the example-based measure and the micro-average measure under different thresholds, and the results are summarized in Table \ref{table:5}. Our goal is to obtain a maximized MiF.

\end{document}